# Learning Contextualized Music Semantics from Tags via a Siamese Neural Network

UBAI SANDOUK and KE CHEN, The University of Manchester

Music information retrieval faces a challenge in modeling contextualized musical concepts formulated by a set of co-occurring tags. In this paper, we investigate the suitability of our recently proposed approach based on a Siamese neural network in fighting off this challenge. By means of tag features and probabilistic topic models, the network captures contextualized semantics from tags via unsupervised learning. This leads to a distributed semantics space and a potential solution to the out of vocabulary problem which has yet to be sufficiently addressed. We explore the nature of the resultant music-based semantics and address computational needs. We conduct experiments on three public music tag collections –namely, CAL500, MagTag5K and Million Song Dataset– and compare our approach to a number of state-of-the-art semantics learning approaches. Comparative results suggest that this approach outperforms previous approaches in terms of semantic priming and music tag completion.

## 1. INTRODUCTION

Music information retrieval (MIR) is becoming increasingly necessary with the rising rates of music production and demand [Serra et al. 2013]. Manual classification of music is infeasible due to costs, biases and contradictions introduced by individual experts. As a result, automatic music understanding is vital for providing viable services. Unfortunately, there exists a gap between what machines extract from music and the corresponding human-level understanding, which is well known as the *semantic gap* [Smeulders et al. 2000]. Efforts to bridge this gap include engineering features [Lew et al. 2006], modeling users' behavior [Schedl et al. 2013] and using concept semantics. Aside from introducing expert knowledge as ontologies [Kim et al. 2008], semantics learning has been dominated by music annotations or tags [Bertin-Mahieux et al. 2010]. Music tracks are associated with textual tags to convey human interpretable concepts describing these tracks. Music understanding is often reduced to automatic annotation of music with suitable tags. Thanks to crowd-sourced [Turnbull et al. 2008] and game-based tagging [Law et al. 2007], large collections of tagging information are accessible to music research. The overall quality of the learnt semantics is determined by the quality of the annotation and how they are used.

Traditionally, mapping music to tags is realized via multi-label classification. For each tag, a dedicated binary classifier is used in order to predict its appropriateness to a track [Turnbull et al. 2007]. This approach loses tag correlation information concerning much of the intention of any tag's use [Sordo 2011, p.34]. Additionally, this approach depends on analyzing the musical content which may vary radically when conveying similar concepts. It is particularly difficult to extend this approach to different MIR tasks and applications in the presence of out of vocabulary (OOV) tags. OOV tags refer to those tags that appear in application but are not observed during semantics learning. Generalizing semantics to new tasks as well as new tags is vital for music understanding. Given that the majority of the musical concepts can be represented in tag form, we believe that proper semantics can be learnt solely from the analysis of co-occurring tags describing music tracks. In this case, intention and meaning of tags are not affected by the musical content, manually defined ontologies and dictionaries or any other information source.

Semanitcs analysis of tag collections can be done in different ways. Levy and Sandler [Levy and Sandler 2008] suggest the use of semantically coherent methods to obtain tag representations, e.g., Latent Semantic Indexing (LSI) [Deerwester et al. 1990] and Probabilistic Latent Semantic Analysis (PLSA) [Hofmann 1999]. These methods uncover statistics that govern the collective use of tags. LSI produces one unique representation for each tag regardless of how it is used in different tracks. We refer to this type of semantics as *global relatedness* semantics. Another global relatedness model is obtained via analyzing the tag collection using aggregation

[Markines et al. 2009] followed by Principle Component Analysis (PCA). Conversely, PLSA produces coherent representations for groups of tags and produces multiple representations for each tag encoded within a representation of a group of tags. We refer to this type of semantics as *contextual relatedness*. Another contextual relatedness model is obtained via Latent Dirichlet Allocation (LDA) [Blei et al. 2003], which produces a compact representation of a group of tags encoded as activation levels of a set of latent *topics*. Alternatively, each group of tags can be represented as a vector of binary tag-relevance indicators known as Bag-of-Words (BoW) [Harris 1954]. The BoW can be smoothed by means of Conditional Restricted Boltzmann Machines (CRBM), which results in a tag-based track representation. The smoothed representation captures correlations among tags and is also contextualized by the CRBM's condition information [Mandel et al. 2011], where the condition information is the one-hot representation of the training document carried out by one 'activated' unit corresponding to the ID of the document being used for training. Evidently, such condition information cannot be applied to new documents in application.

While the aforementioned methods are able to capture contextual track-to-tag relatedness, they struggle in capturing underlying tag-to-tag relatedness; leading to inconsistent semantics and difficulty across different MIR tasks. For instance, LDA provides a method for relating topics to tags but does not have a clear measure on relatedness among tags. Additionally, these approaches assume a close-set scenario; i.e., all tags have to be known during training and used according to their predefined meanings. In reality, different tags may be used to mean the same musical concept, e.g., 'drums'/'drumset'; and the same tag may be used to describe more than one musical concept, e.g. 'guitar' to mean different guitar types. In general, the intended meaning of a single tag cannot be revealed unless all other tags used with it for describing a track are examined. Hereinafter, we use *companion tags* to refer to these co-occurring tags in an annotation of a track. Furthermore, new tags beyond a predefined vocabulary may also be used by annotators, i.e., OOV tags. In summary, the existing models fail to yield the proper *contextualized relatedness* semantics between individual tags and, in particular, there appear OOV tags in applications.

In contrast to the slow progress in tag semantics modeling, natural language processing benefits from a class of distributed language models capturing underlying relatedness among words [Mikolov et al. 2013]. Thanks to their simplicity and capacity in providing generic semantics for numerous linguistic tasks, such models yield semantic spaces where words are embedded based on their syntactic similarity. A distance measure is often used within the semantic space to reflect *syntactic relatedness* of word pairs. Unlike the dictionary-based models [Iacobacci et al. 2015], distributed language models are trained using words in a corpus without considering any manually constructed information sources, e.g., dictionary. Such success inspires further research into distributed semantics. Unfortunately, applying linguistic models directly to tag collections is unreasonable due to the lack of syntactic structure among tags. Without the ordering information, these models can only capture global relatedness neglecting contextualized meanings underlying tags uses. Furthermore, these linguistic models falsely assume linguistic properties of tags as same as those of words. However, tags are beyond words as music tags may be symbols, abbreviations and phrases, e.g., "r'n'b", "80s" and "rhythmic loops".

To facilitate our presentation, a *document* hereinafter refers to a set of co-occurring tags, i.e., $\delta = \{\tau_i\}_{i=1}^{m}$, used to describe a music track. For any tag $\tau_i$, $\delta$ forms its local context used to disambiguate the intention of using $\tau_i$ for that track. As a result, the specific meaning of a tag, dubbed a *concept,* is only identified after examining the tag and its local context. In our recent work, we formulate the *concept embedding,* $\boldsymbol{CE}(\tau_i, \delta)$, problem such that the distance between different concepts embedding reflects contextualized relatedness [Sandouk and Chen 2016]. To solve this problem, we proposed a contextualized semantic learning approach by means of Siamese architecture [Bromley et al. 1993]. By using unsupervised learning, the Siamese neural network establishes a semantics space that embeds concepts properly reflecting their relatedness as Euclidean distance. The space contains multiple representations for each tag in different contexts so that it co-locates with other tags that share the same intention and can estimate the concept underlying an OOV tag from its local context. In this paper, we investigate the suitability of this approach in learning musical semantics based on music tag collections. The main contributions in this work are summarized:

— We thoroughly investigate the suitability of Siamese CE approach [Sandouk and Chen 2016] in modeling the musical concepts, including the proper capacity of the network in learning musical semantics, highlighting the emergent structure of the musical tag space and its smooth nature, and assessing the computational efficiency of the network in music domain.
— To make a state-of-the-art contextualized semantic learning model comparable to ours, we propose an improved version for the CRBM model [Mandel et al. 2011] so that it can capture statistical co-occurrence likelihoods of tags as our Siamese architecture does. The semantics learned by the improved CRBM model is significantly different from that done by its original version.
— As the learned CE model can be generalized to tags never seen during training, we examine this non-trivial issue by applying it to the million song dataset, a large dataset facing the long tail problem [Bertin-mahieux et al. 2011].
— We conduct a thorough evaluation on the learnt contextualized semantics based on two benchmark MIR tasks, i.e., semantic priming and tag completion, via a comparative study to several state-of-the-art semantic learning models.

The paper is organized as follows: Section 2 reviews the CE model and achieving concepts embedding for tags. Sections 3 and 4 report experimental results in the semantic priming and the tag completion tasks, respectively. Section 5 discusses relevant issues while Section 6 concludes with future works related to this research, including possible concrete uses of CE semantics in MIR tasks.

## 2. MODEL DESCRIPTION

For self-containment, we review our recently proposed approach for learning contextualized semantics from tags [Sandouk and Chen 2016]. As the underpinning techniques, this model is applied to learning semantics from musical tags.

### 2.1 Tag and Context Features

Tags are represented via aggregation [Markines et al. 2009] which captures their global relatedness over the entire training set. Each tag use is weighted using $tfidf$ [Singhal 2001]. Given one training tag $\tau \in \Gamma$ and one training document $\delta \in \Delta$ where $\Gamma$ is the vocabulary and $\Delta$ is the document collection: their binary relatedness is given

$$tf(\tau, \delta) = \begin{cases} 1 & \text{when } \tau \text{ appears in } \delta \\ 0 & \text{otherwise} \end{cases}.$$

The rarity of a tag is considered using inverted document frequency

$$idf(\tau) = log\left(\frac{|\Delta|}{1 + |\{\delta \in \Delta : tf(\tau, \delta) = 1\}|}\right), \text{ where } |.| \text{ is the cardinality of a set.}$$

The $tfidf$ weight is defined by the product of the binary relatedness and the inverted document frequency as $tfidf(\tau, \delta) = tf(\tau, \delta) \times idf(\tau)$.

Finally, each tag is described by its usage pattern across all training documents as $\boldsymbol{u}(\tau) = \{tfidf(\tau, \delta_i)\}_{i=1}^{|\Delta|}$. Consequently, global relatedness between two training tags $\tau_1, \tau_2 \in \Gamma$ is measured by the dot product of their respective usage vectors $T(\tau_1, \tau_2) = <\boldsymbol{u}(\tau_1), \boldsymbol{u}(\tau_2)>$, which leads to tag representation of $|\Gamma|$ features as

$$t(\tau) = \{T(\tau, \tau_i)\}_{i=1}^{|\Gamma|}. \quad (1)$$

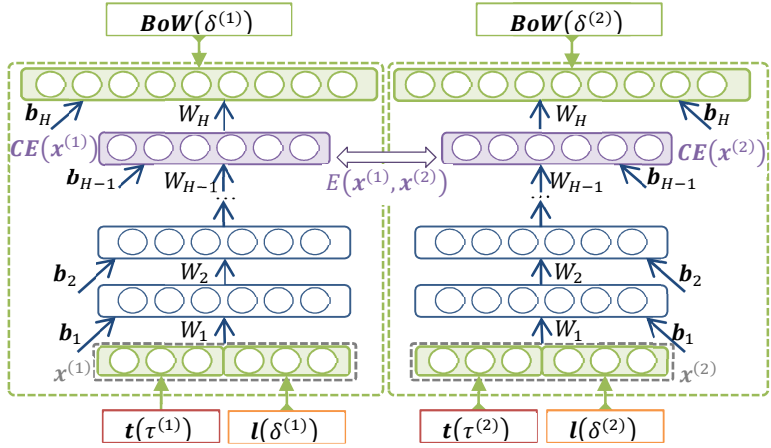

Figure 1: The Siamese network for contextualized music semantics learning (see Section 2.2 for notation).

Local contexts are captured via LDA [Blei et al. 2003] over all tags in a document. LDA assumes a set of latent independent topics $\Phi$ that softly cluster the documents based on the used tags resulting in a probabilistic topic model (PTM) representation of the entire document. During training, the process estimates scalar priors B for the Dirichlet distributions to model the tags within each topic as well as the scalar prior $B^0$ in the Dirichlet distribution used to model the topics. After training, the probability of a tag $\tau \in \Gamma$ is subject to $p(\tau|\phi) \sim Categorical(Dirichlet(B))$ given a topic $\phi \in \Phi$ where $p(\phi) \sim Dirichlet(B^0)$. For a document $\delta$, $p(\phi|\delta) \sim p(\phi)\prod_{\tau \in \delta} p(\tau|\phi)$, the local context of a tag is represented by a vector of $|\Phi|$ features corresponding to the $|\Phi|$ topic distribution output with respect to $\delta$:

$$l(\delta) = \{l_c(\delta)\}_{c=1}^{|\Phi|}, \; l_c(\delta) = p(\phi_c|\delta). \quad (2)$$

### 2.2 Siamese Architecture

To learn contextualized semantics, we use the Siamese architecture illustrated in Figure 1. We train a deep neural network that predicts companion tags for a tag based on the input features. Given an example $x_k$, consisting of tag $\tau$ and document $\delta$, input features are the result of concatenating the tag and context features: $x_k(\tau, \delta) = \{t(\tau), l(\delta)\}$. The network consists of $H$ hidden layers where a layer $h$ is characterized by weight $W_h$ and bias $b_h$ parameters. The output of layer $h$ for example $x_k$ is

$$z_h(x_k) = f(W_h \cdot z_{h-1}(x_k) + b_h), 1 \le h \le H,$$

where $f(.)$ is the element-wise hyperbolic tangent function. The output of the $(H-1)^{th}$ hidden layer, i.e., the penultimate layer, is used as the semantic representation. We refer to this representation as *contextual embedding* (CE). We stipulate the raw input features are $x = z_0(x)$, the contextualized embedding is $CE(x) = z_{H-1}(x)$ and the prediction is $\hat{y}(x) = z_H(x)$. Lastly, for a pair of inputs to the Siamese network $x^{(1)} = \{t(\tau^{(1)}), l(\delta^{(1)})\}$ and $x^{(2)} = \{t(\tau^{(2)}), l(\delta^{(2)})\}$, the embedding similarity is defined by the Euclidean distance between their contextual embedding:

$$E(x^{(1)}, x^{(2)}) = \|CE(x^{(1)}) - CE(x^{(2)})\|_2.$$

Hereinafter, we shall drop all explicit parameters to simplify presentation, e.g., $y_k$ stands for $y_k(x_k(\tau, \delta))$ and $y_{kj}$ means the $j^{th}$ entry of $y_k$.

### 2.3 Learning Algorithm

For each training document $\delta$ of $m$ tags, we create $m$ *positive* training examples where each example consists of one focused tag and the shared local context features. The prediction targets for these examples are the BoW of $\delta$. Moreover, we artificially synthesize $m$ *negative* training examples where each example consists of one focused tag that does not appear in $\delta$ and the local context features of $\delta$. The prediction targets for these examples are the complement of BoW of $\delta$. These examples reduce the domination of context features in predicting the related tags.

The deep network is pre-trained with the greedy layer-wise initialization with sparse autoencoders suggested in [Bengio et al. 2007]. The initialization is followed by the error back propagation training to predict the target representation from the tag and context input. The loss is a variant of the cross-entropy cost. Given the entire training dataset $X = \{x_k(\tau, \delta)\}_{k=1}^{K}$, the loss is thus defined by

$$\mathcal{L}_P(X; \Theta) = -\frac{1}{2K|\Gamma|} \sum_{k=1}^{K} \sum_{j=1}^{|\Gamma|} \left( \kappa_k (1 + y_{kj}) \log(1 + \hat{y}_{kj}) + (1 - \kappa_k)(1 - y_{kj}) \log(1 - \hat{y}_{kj}) \right). \quad (3)$$

Here $\kappa_k = \frac{\left|\{i: y_{ki}=1\}_{i=1}^{|\Gamma|}\right|}{|\Gamma|}$ is a weight that highlights the cost of a false negative error. This optimization is carried out via the Stochastic Back Propagation (SBP) [Bottou 2012]. After completing the learning of this network, we obtain the *initial embedding*.

We train the Siamese architecture by coupling two copies of the initial network operating together for distance learning. Given an input examples pair $\boldsymbol{x}^{(1)}$ and $\boldsymbol{x}^{(2)}$, the contextualized semantic similarity is reflected by the Kullback–Leibler (KL) divergence of their respective contexts:

$$KL(\boldsymbol{x}^{(1)}, \boldsymbol{x}^{(2)}) = \sum_{c=1}^{|\Phi|} \left( \left(l_c^{(1)} - l_c^{(2)}\right) \log\left(\frac{l_c^{(1)}}{l_c^{(2)}}\right) \right).$$

Moreover, the distance learning makes use of the semantic similarity to penalize the embedding if distances deviate away from a target distance. Given a pair of input examples $\boldsymbol{x}_n^{(1)}$ and $\boldsymbol{x}_n^{(2)}$, we stipulate $\mathbb{E} = E(\boldsymbol{x}_n^{(1)}, \boldsymbol{x}_n^{(2)})$ and $\mathbb{S} = e^{\frac{-\lambda}{2}\left(KL(x_n^{(1)}, x_n^{(2)})\right)}$ where $\lambda$ is a sensitivity parameter that affects the degree to which the embedding is dominated by context similarity. Given two subsets $X^{(1)}$ and $X^{(2)}$ of cardinality $N$ of randomly selected examples via pairing from $X$, the following loss is used for Siamese learning:

$$\mathcal{L}_S(X^{(1)}, X^{(2)}; \Theta) = \frac{1}{N} \sum_{n=1}^{N} \left( I_1 \left(\mathbb{E} - \beta(1 - \mathbb{S})\right)^2 + I_2 \rho \left(\mathbb{E} - \beta(1 - \mathbb{S})\right)^2 + I_3 (\mathbb{E} - \beta)^2 \mathbb{S} \right), \quad (4)$$

where $\sum_{k=1}^{3} I_k = 1; I_k \in \{0,1\}$ are indicators which distinguish three possible cases of input pairs: $I_1 = 1$ when both inputs are positive examples; $I_2 = 1$ when both inputs are negative examples; and $I_3 = 1$ when one input is a positive example and the other is a negative one. $\beta$ is a scaling factor used to ensure tags spread over the embedding space and $0 \leq \rho < 1$ is an importance factor which reduces the effects of case $I_2 = 1$.

During distance learning, each component network is also trained simultaneously to predict the BoW of its input to avoid an abrupt change of the CE learnt initially with a single network. Thus, the distance learning needs to combine two loss functions defined in Equations 3 and 4, which leads to a multi-objective loss:

$$\mathcal{L}(X^{(1)}, X^{(2)}; \Theta) = \sum_{i=1}^{2} \mathcal{L}_P(X^{(i)}; \Theta) + \alpha \mathcal{L}_S(X^{(1)}, X^{(2)}; \Theta), \quad (5)$$

where α is a trade-off parameter that reconciles two different losses.

The optimization problem defined in Equation 5 is solved via SBP. Iteratively, small batches of examples are randomly selected, their loss measured and the parameters updated accordingly. The networks are kept identical by averaging the parameters after each update. Training continues until validation $P@2$ (c.f. Section 3.5.) stops improving. See the appendix of [Sandouk and Chen 2016] for further details on learning algorithm.

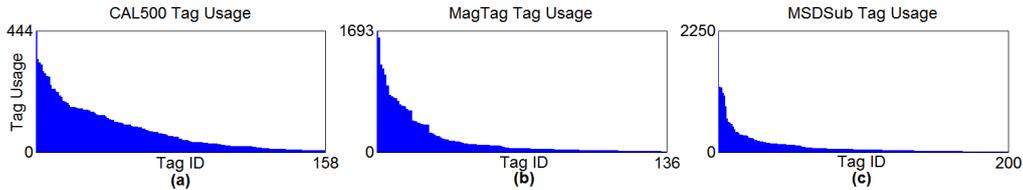

Figure 2: Frequency of tags used in different datasets. (a) CAL500. (b) MagTag5K. (c) MSDSub.

### 2.4 Tag Contextual Embedding

For an input tag $x(\tau, \delta) = \{t(\tau), l(\delta)\}$, an embedding network is used to obtain the CE representation denoted by $CE(\tau, \delta)$. For the two-stage learning procedure, the model generates two $CE(\tau, \delta)$ representations by using the network trained with only the prediction loss in Equation 3 or using the multi-objective loss in Equation 5.

For an OOV tag $\tau_{oov}$ appearing in a document $\delta$ alongside in-vocabulary tags, i.e., $\delta = \tau_{oov} \cap \delta_{iv}; \delta_{iv} = \{\tau_i\}_{i=1}^{|\delta_{iv}|}$, we use CE representations of all $|\delta_{iv}|$ companion tags denoted as $\{CE(\tau_i, \delta_{iv})\}_{i=1}^{|\delta_{iv}|}$ and estimate an OOV tag representation as the centroid of these vectors $CE(\tau_{oov}, \delta) = \frac{1}{|\delta_{iv}|}\sum_{i=1}^{|\delta_{iv}|} CE(\tau_i, \delta_{iv})$. This CE representation allows the estimation of intentions of OOV tags based on local contexts.

## 3. EVALUATION OF LEARNT SEMANTICS

In this section, we first describe the datasets and the settings used in our experiments. Then we demonstrate the contextualized semantics learnt by the CE model via visualization. Finally, we report experimental results on semantic priming.

### 3.1 Dataset and Feature Extraction

In our experiments, we employ three publically available datasets: CAL500 [Turnbull et al. 2007], MagTag5K [Marques et al. 2011] and Million Song Dataset (MSD) [Bertin-mahieux et al. 2011]. These datasets exhibit different tag usage distributions as shown in Figure 2. In tagging, there exists the so-called long-tail problem where the majority of available tags are rarely used. From Figure 2, it is observed that CAL500 does not suffer from this problem as severely as the other two datasets.

CAL500 is a dataset of 500 songs annotated using 158 unique tags via surveys. The dataset uses tags densely given the fact that there are 25 tags per document on average. MagTag5K is a controlled version of MagnaTune where repeats and contradictions have been removed. MagnaTune is the result of an online annotation game that allowed users to evaluate the appropriateness of complete tag sets rather than individual tags at a time [Law et al. 2009]. MagTag5K contains 5,259 documents and a vocabulary of 136 tags. It is sparser than CAL500 with five tags per document on average. MSD is a dataset of one million songs' information. Many of the songs are tagged through last.fm, a crowd sourced annotation website. The original dataset contains a vocabulary of 520,539 tags. In our work, we use the 300 most used tags and the 14,627 documents using only these tags to form a subset, named *MSDSub*. MSDSub has 3.2 tags per document on average.

We simulate OOV scenarios by reserving a number of tags away from semantics learning; once a tag is reserved, any relevant documents containing it is never used in training the model. We randomly select and reserve 22 tags and the 1,160 relevant documents in MagTag5K and 100 tags and 7,054 relevant documents in MSDSub. The remaining documents are used for feature extraction where we obtain 114 and 200 tag features for MagTag5K and MSDSub, respectively. However, the high density of CAL500 does not allow for such setting since reserving a single tag would disable around 160 relevant documents while there are only 500 documents in this dataset. Thus, all 158 tags are used in training, which yields 158 tag features. To model local contexts, we empirically sought the proper number of topics in LDA by using the Dirichlet Hierarchical Process suggested in [Teh et al. 2006]. As a result, 25, 19 and 23 topics are used for CAL500, MagTag5K and MSDSub, respectively.

Table I. Averaging time (second) taken in different training stages and test.

| Dataset | CAL500 | MagTag5K | MSDSub |
|---|---|---|---|
| Pre-training (/fold) | 190 | 134 | 177 |
| Prediction training (/fold) | 6149 | 12406 | 20495 |
| Siamese training (/fold) | 3032 | 3068 | 3928 |
| CE representation extraction (/instance) | $2.3 \times 10^{-3}$ | $1.5 \times 10^{-3}$ | $3.3 \times 10^{-3}$ |

### 3.2 Experimental Settings

In our experiments, we adopt three-fold cross-validation (CV) by randomly splitting a dataset into three subsets of equal sizes. In each fold, two thirds are used for training the CE model and the remaining third is used for validation and testing. As a result, in each fold, there are 40, 300 and 500 documents for validation and 127, 1,007 and 3,487 documents for test in CAL500, MagTag5K and MSDSub, respectively. Determining the hyper-parameters is done via grid search. It is worth stating that the initialization of the deep network is done with sparse auto-encoders in a greedy layer-wise way instead of a random initialization. Consequently, we found a proper network structure: *input→100→100→10→output* where $\lambda = 0.5, \beta = 3$ and $\alpha$ values are in the range $[0.1, 1]$ for all three datasets.

For thorough evaluation, the CE model is evaluated in two forms: the trained initial semantics (CE) and the fine-tuned Siamese semantics (Siamese CE). Furthermore, we compare our approach to several state-of-the-art approaches reviewed in Section 1, including PLSA [Levy and Sandler 2008], LDA [Law et al. 2010] and CRBM [Mandel et al. 2011]. Unfortunately, CRBM uses a one-hot representation for condition or context. While this condition produces smoothed BoW representations for all the training documents, learnt semantics cannot be applied to new documents and across MIR tasks. For proper comparison between CE and CRBM, we come up with an improved version for CRBM by replacing the one-hot condition representation with the same context representation used in CE, i.e., PTM. We name this non-trivial extension *CRBM(PTM)*. The architecture and cost function in [Mandel et al. 2011] remain the same in CRBM(PTM). It is worth highlighting that the semantics learnt with CRBM(PTM) are significantly different from those learnt by the original CRBM and the biggest difference appears in those documents of similar PTM context features having maximally dissimilar one-hot context representations. The CRBM model implicitly encodes the "popularity" of tags in its bias vector so that it is more likely to predict popular tags. Among all the models used for comparison, the resultant semantics of CRBM(PTM) are the closest to those obtained from the CE model. We also include the Random model which responds with random relatedness values between pairs of tags. All the results reported in this section are based on the *test* subset in three folds. Such results indicate the generalization ability of the learnt semantics. Apart from the goal from a learning perspective, results on training data are also meaningful and can be directly employed in various IR tasks. Due to the limited space, however, we have to report results on training sets in an appendix.

### 3.3 Computational Efficiency

It is well known that training deep neural networks often takes long times and may need the use of GPUs to speed up the training [Raina et al. 2009]. In our experiments, we use a Linux server with 24 CPUs running at 2.0 GHz and memory of 128GB. All our algorithms were implemented in Matlab® 2012a.

For three-fold CV, the average training time at different training stages is listed in Table I. Also the average time in extracting the CE representation of an instance in test is reported. Obviously, the training time generally depends on the number of input features and the size of different datasets. In particular, we observe that the pre-training on CAL500, a dataset of densely used tags, takes longer than that on MagTag5K and MSDSub. Time spent during test for CE representation extraction is similar in the three datasets despite the difference in the number of input features.

### 3.4 Visualization

By using unsupervised t-SNE [van der Maaten and Hinton 2008], we project the CE representations of all tags used in 200 randomly selected documents from MagTag5K onto a 2D plane for visualization. As illustrated in Figure 3(a), each dot represents a tag in one document. To see the space clearly, we annotate dots with their corresponding tags. However, the annotations render the plot overly crowded and difficult to read. Therefore, we highlight only four densely populated regions confined in these boxes in Figure 3(a) and illustrate the details of four regions in Figures 3(b)-3(e) by zooming in on four regions. Within these plots, close duplicates of any tag were annotated only once for better visual effect. Moreover, we label each region with the captured concepts according to human knowledge. It is observed that four regions in Figures 3(b)-3(e) correspond to the concepts of "Pop and Dance", "New Age and Ambient", "Types of Singing" and "Classical Music", respectively.

From Figures 3(a)-3(e), it is observed that there is a swift transition between different musical concepts and the continuous nature of the embedding space is noticeable. For example, there is a clear transition in Figure 3(c) from electrical new age music (up left) to acoustic classical ambient (down right). In Figure 3(e), there is a clear transition from concepts associated with piano-based classical music (down right) through baroque music (middle center) to folk classical music (up left). In addition, transitions between the different regions are swift as well, e.g., those tags concerning Rock music (loud, electric, rock, etc…) near the right boundary of the "Pop and Dance" region in Figure 3(b) and those near the left boundary of the "Types of Singing" region in Figure 3(d). A similar effect is also observed between the "Pop and Dance" and the "New Age and Ambient" boundaries in Figures 3(b) and 3(c).

As stated previously, a complete musical concept is often formulated by using a set of coherent tags collectively instead of an individual tag. This has been carried out by co-locating such coherent tags in CE space. For instance, tags "eerie", "scary", "deep" and "ambient" seen on the middle left in Figure 3(c) collectively form a single musical concept corresponding to a specific type of chilling ambient music. Similarly, "harp", "no.singing" and "piano" at the bottom center of Figure 3(e) collectively form another musical concept describing a specific type of classical music. We also observe that many tags are located in multiple regions in order to describe different musical concepts in context. For instance, the tag "slow" has been found simultaneously in Figures 3(c), 3(d) and 3(e) to express slow and quiet types of ambient sounds and music relevant to the "New Age and Ambient", slow songs and singing voice linked to the "Types of Singing" and slow classical music often correlated with the use of violins and flutes pertaining to "Classical Music", respectively.

Figure 3(f) shows a 2D projection of CE representations of all 388 instances of the "guitar" tag found in MagTag5K, where three clusters emerge. According to our genre knowledge, we identify that two clusters marked by "■" are associated with the documents of acoustic nature and the third one marked by "□" corresponds to the documents of electric nature. This visualization clearly demonstrates the polysemous aspects of a tag in different contexts. Here, we emphasize that our learning model captures different meanings of a tag based on the intention of using this tag rather than mere co-occurrence with other tags.

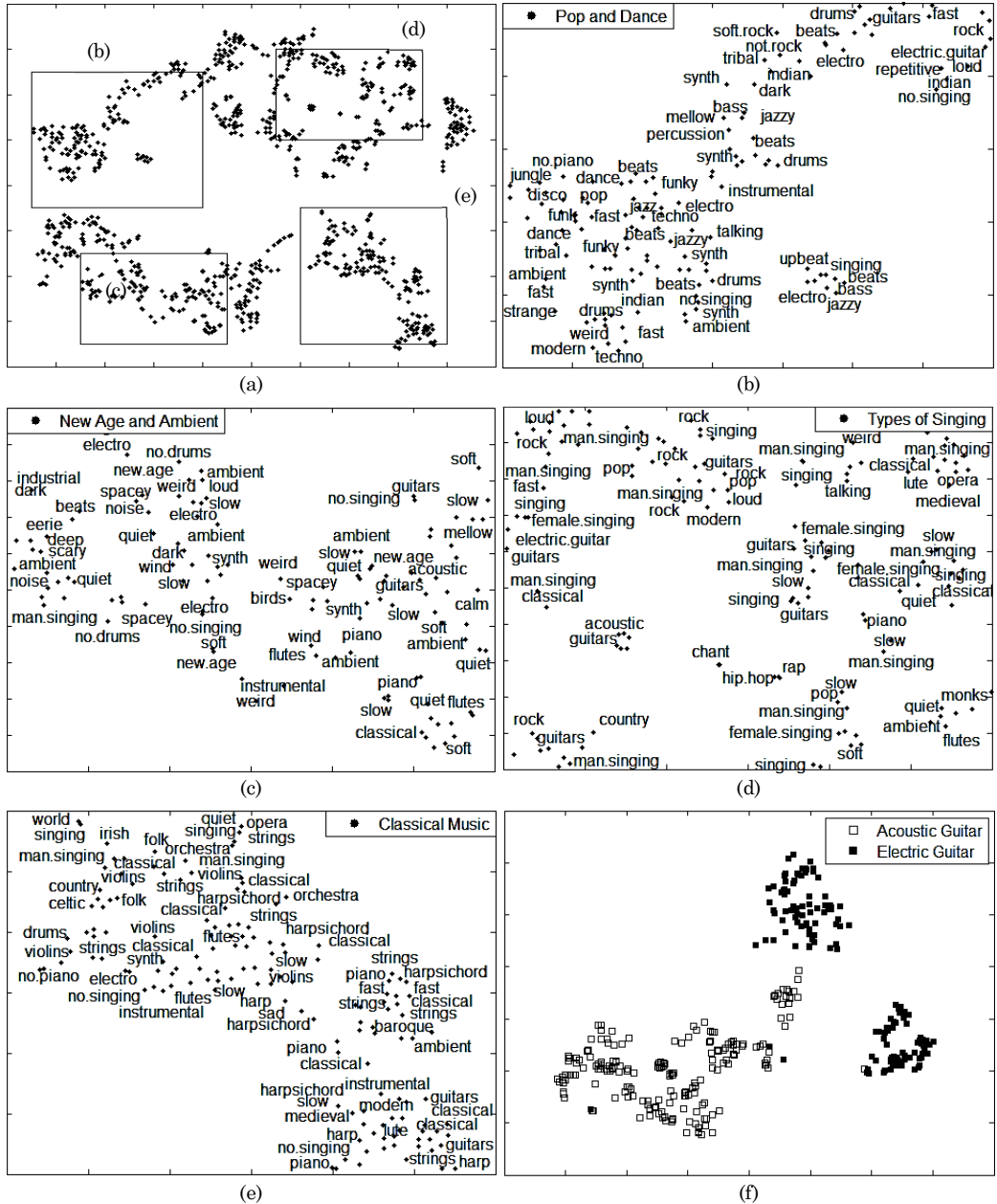

Figure 3: 2D projection of CE representations of exemplar tags. (a) All tags in 200 randomly selected documents form MagTag5K. (b)-(e) Annotations of tags inside four boxed regions in (a). (f) All 388 "guitar" tag instances in MagTag5K.

In summary, the visualization in Figure 3 demonstrates several useful properties that facilitate MIR, including co-located musical concepts, semantic distance within the CE space reflecting the corresponding contextualized relatedness, swift transition between musical concepts and the polysemous aspects of a tag in different contexts.

### 3.5 Semantic Priming

For evaluation of learnt semantics, we use *semantic priming*, a benchmark IR task, where all semantically related tags are expected to be identified given a query tag [Lund and Burgess 1996]. Successful priming is observed when the model is presented with a query tag in context and the model is able to identify related tags without including any non-related ones. This abstract task acts as a generic natural evaluation test bed for semantics without favoring any application. In order to complete the evaluation, we need a set of query tags in context along with their ground truth information that associates related tags under that context. However, due to the lack of such ground truth information, we iteratively use one evaluation document as a coherent set of semantically related tags that should be identified as a response to a query tag. As a result, all tags in an evaluation document $\delta$ are used as ground truth and each of these tags is used as a query tag in turn. In other words, the quality of semantics learnt by different models can be evaluated by examining how well one query tag and its local context can identify its companion tags in the same evaluation document. For such a task, a learning model is assessed by measuring the performance of retrieving the top $k$ tags in response to a query tag. Given $r$, the predicted ordered list of tags, and $\bar{r}_k$, the top $k$ tags of $r$, *Precision at $k$* is

$$P@k(k; \delta, r) = \frac{|\delta \cap \bar{r}_k|}{k},$$

which indicates the accuracy of the model in correctly identifying $k$ related tags given one query tag. Unfortunately, this measure is affected by the length of an evaluation document such that its values drop quickly when $k$ exceeds the length of the evaluation document and hence evaluation across multiple documents becomes meaningless. This issue can be solved using *Mean Average Precision* (MAP): the average $P@k$ values for a query up to the number of tags in the evaluation document

$$MAP = \frac{1}{|\delta|} \sum_{i=1}^{|\delta|} P@k(i; \delta, r).$$

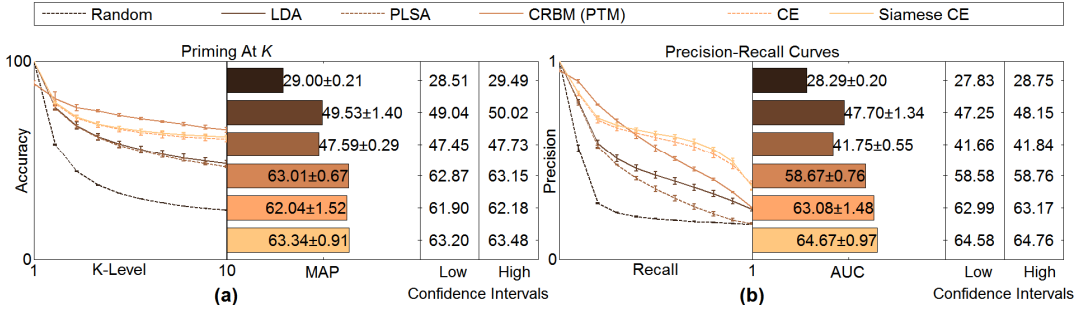

Figure 4: Semantic priming performance of different models on test subsets of CAL500. (a) The priming accuracy at different $k$ levels as well the MAP mean and standard errors. (b) The Precision-Recall curves as well as the AUC mean and standard errors. This notation is applied to all the figures hereinafter.

Intuitively, MAP measures the percentage of the identified tags that are "correct". High MAP results mean that the assessed model identifies relevant tags at the top of the retrieved list or in the low recall range. However, it does not measure the performance over the entire retrieved list. In fact, the performance in the high recall range is often important for tasks such as tag completion and query expansion where all the related tags need to be identified. Therefore, we also evaluate the performance over the entire retrieved list by measuring the numbers of retrieved tags required to achieve the standard 11 recall levels: 0.0, 0.1, 0.2, …, 1.0 and their corresponding precision [Manning et al. 2009] with $Recall(\delta, \bar{r}) = \frac{|\delta \cap \bar{r}|}{|\delta|}$ and $Precision(\delta, \bar{r}) = \frac{|\delta \cap \bar{r}|}{|\bar{r}|}$. Finally, we aggregate all the results in a single figure of merit by using the *Area Under Curve* (AUC). Unlike the *Receiver Operating Characteristic* (ROC) for binary classification, our used AUC allows the assessment of a ranked list. For two models with similar MAP levels, higher AUC suggests better performance in the high recall range, i.e., identifying *all* the relevant tags. Figures 4, 5 and 6 depict the MAP and the AUC results for different contextualized semantic learning models to be compared on three datasets, CAL500, MagTag5K and MSDSub, respectively.

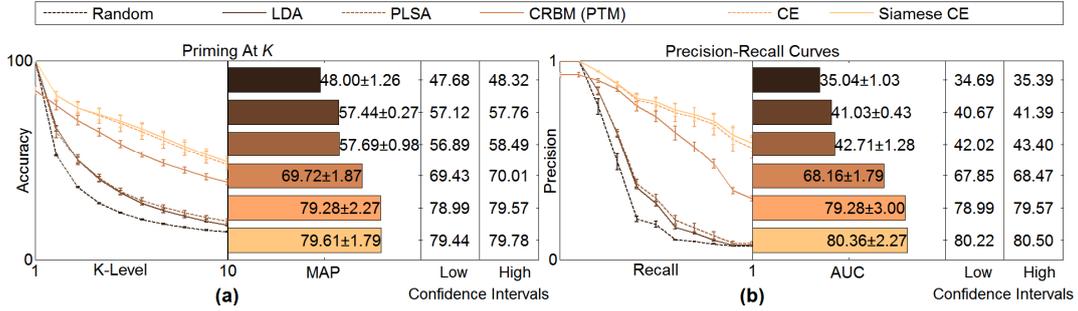

Figure 5: Semantic priming performance of different models on test subsets of MagTag5K.

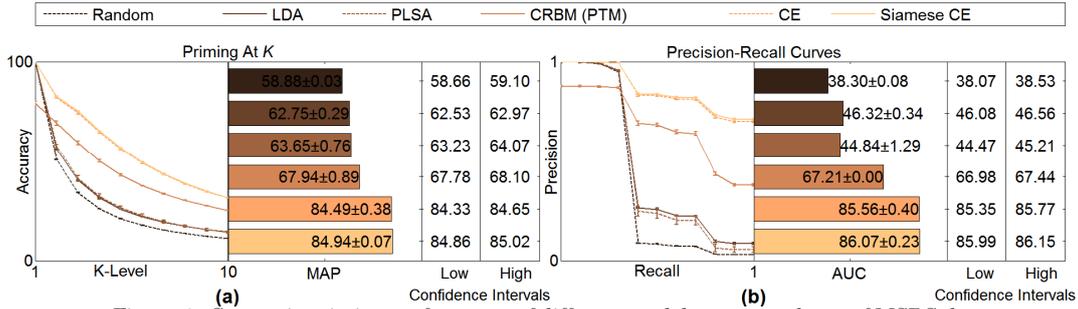

Figure 6: Semantic priming performance of different models on test subsets of MSDSub.

Figure 4 shows the semantic priming results on CAL500. As this is a dense dataset, relevant tags are correctly predicted in small $k$ values by most models. For example, the Random model achieves 56% in $P@2$ shown in Figure 4(a). Moreover, CRBM(PTM) is comparable to Siamese CE in terms of MAP and has a higher precision in the recall range of [0.1,0.3] because it tends to predict popular tags. The good performance at the high recall range of CE is evident in Figure 4(b) and reciprocally results in a clear advantage in AUC. Overall, Siamese CE outperforms other models with statistical significance (Student's t-test p-value < 0.01). Confidence intervals at 95% reveal a gain in accuracy when using Siamese CE as indicated by AUC. Moreover, we measure the standardized difference, i.e., effect size, between the reported statistics. For our purposes, we report Cohen's d effect size which measures the standardized difference between two independent samples means [Cohen 1977, pp.19–74]. Cohen's d effect size between Siamese CE and CRBM(PTM) is 0.42 in MAP and 7.00 in AUC, and the effect size between Siamese CE and CE is 1.04 in MAP and 1.27 in AUC. These results suggest a clear advantage for Siamese CE over CRBM(PTM) in the low recall range as well as a significant advantage in the high recall range as shown in Figure 4(b). In contrast, it is evident from Figure 4 that the performance of LDA and PLSA is inferior to that of CRBM(PTM) and that of the CE model. It is also observed from Figure 4(b) that all models struggle in the high recall range due to having only 335 training documents. The lack of sufficient training data on this dense dataset inevitably limits the generalization ability of any model.

Figure 5 shows the semantic priming results on MagTag5K. In this case, the CE model performs better than all other models especially in the high recall range with statistical significance (Student's t-test p-value < 0.01). Short documents in this dataset account for the high MAP accuracy of the Random model. We observe that LDA and PLSA perform similar to each other with little improvement over the Random model. In contrast, CRBM(PTM) performs better than both LDA and PLSA but less significantly than observed on CAL500 due to the sparsity of MagTag5K. We notice that the Siamese CE outperforms CE as well as reduces variance of results amongst different folds. The confidence intervals reveal a slight advantage by using Siamese CE in MAP and a significant advantage in AUC. Moreover, the effect size

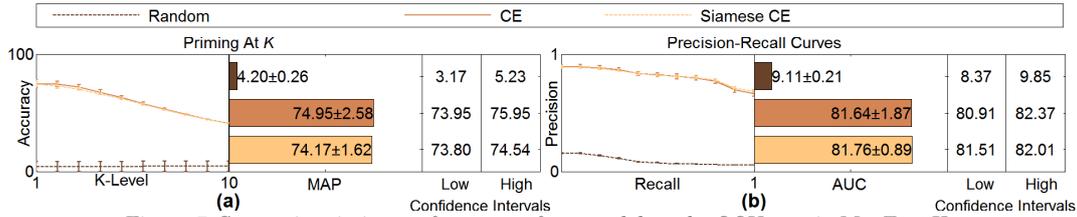
Figure 7: Semantic priming performance of our model on the OOV tags in MagTag5K.

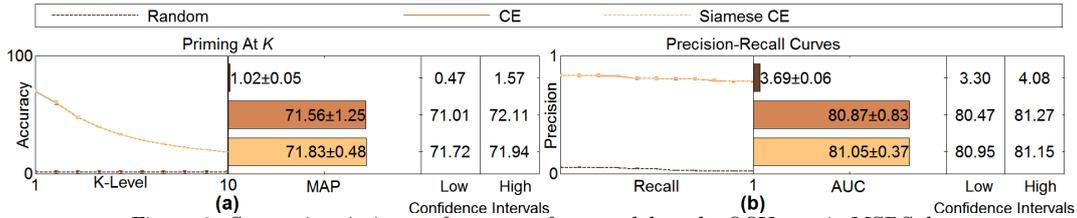
Figure 8: Semantic priming performance of our model on the OOV tags in MSDSub.

between Siamese CE and CRBM(PTM) is 5.49 in MAP and 5.56 in AUC, and the effect size between Siamese CE and CE is 0.16 in MAP and 0.41 in AUC. It is worth clarifying that the advantage of Siamese CE over CE looks modest in terms of generalization but the nature of this dataset, the difficulty in understanding tags and the advantage gained by Siamese CE in the high recall range, as evident in Figure 5(b), suggest that Siamese CE may be more suitable to some MIR applications.

Figure 6 shows the semantic priming results on MSDSub. Due to the extremely short document lengths (3.2 tags on average) in MSDSub, the $P@k$ drops quickly after $k = 3$ for all models. It is seen from Figure 6(a) that both CE and Siamese CE outperform other models with statistical significance (Student's t-test p-value < 0.01). In particular, the CE model performs extremely well in the high recall range while LDA and PLSA perform poorly and similarly to the Random model. The deteriorated performance of other models in the high recall range is due to the size of the vocabulary and the sparsity of this dataset which does not affect the CE model. Although CRBM(PTM) still outperforms LDA and PLSA, it clearly underperforms the CE model as evident in Figure 6. Similar to results on MagTag5K, the Siamese CE slightly outperforms CE and reduces the variance amongst different folds. The effect size between Siamese CE and CRBM(PTM) on MSDSub reaches 26.88 in MAP and 117.64 in AUC, and the effect size between Siamese CE and CE is 1.67 in MAP and 1.54 in AUC. The same conclusion drawn from results in MagTag5K is applicable in this dataset. Hence, the CRBM(PTM), our extension of the CRBM [Mandel et al. 2011], is a good candidate for capturing contextualized relatedness between tags as it always outperforms LDA and PLSA. Nevertheless, the CE model generally yields statistically significant better results than the CRBM(PTM) on all three datasets according to all four evaluation criteria. In particular, the good performance in the high recall range suggests that semantics learnt by the CE model would facilitate auto-annotation or auto-tagging, tag completion and semantic query expansion required by MIR tasks.

The OOV problem is not unique to the CE model. It is encountered whenever the semantics is used in applications where tags' use is not restricted; such as, online tagging services which allow users to tag music using any tag no matter if such tag had been used before. Although the concept conveyed in an OOV tag is less certain than an in-vocabulary tag, it is very important to exploit semantics underlying the OOV tag. To the best of our knowledge, none of the existing semantic learning models address the OOV issue. Fortunately, we can infer the intention of an OOV tag based on the CE representations of co-occurring companion tags in the document containing the OOV tag. In our experiments, OOV tags are used as query tags for semantic priming and then the primed lists are compared against their ground truth in-vocabulary tags in the evaluation document. To demonstrate the effectiveness of our approach, we also applied the Random model for baseline performance.

Figures 7 and 8 show the priming results on the OOV tags reserved in MagTag5K and MSDSub (c.f. Section 3.1). It is observed that Siamese CE performs slightly better than CE due to the learnt semantic distance reflecting the contextualized relatedness better. By comparing to results on test subsets of in-vocabulary tags shown in Figures 5 and 6, the OOV performance of the CE model is quite close on two datasets while the Random model completely fails. By using the Cohen's d effect size, the difference between Siamese CE and CE is -0.36 in MAP and 0.08 in AUC for in MagTag5K as well as 0.29 in MAP and 0.28 in AUC for MSDSub. In general, the consistent performance has been observed for both datasets.

In summary, the semantics learnt by the CE model considerably outperforms that obtained by other state-of-the-art models (even with a non-trivial extension). In particular, the CE model is good at capturing semantics in the sparse BoW scenario and has the unique capability of inferring the semantics from the OOV tags. Hence, we firmly believe that the contextualized semantics learnt by the CE model would facilitate various MIR tasks.

### 4. TAG COMPLETION

In this section, we demonstrate the effectiveness of the semantics learnt by the CE model in *tag completion*, a benchmark MIR task that requires suggesting complementary tags to an existing group of tags describing a music track. Likewise, we also compare the CE model to those models used in the semantic priming evaluation and report comparative results in the tag completion task.

Unlike semantic priming where a query concept is used to identify only its related tags, a query concept in tag completion would result in a score for *all* tags in descending order in terms of semantic relatedness. Performance evaluation in this task requires the continuous relatedness ground truth that properly reflects semantic coherence among all tags in different contexts. Such information is neither required nor available in the semantic priming task described in Section 3.5. This new information used for evaluation properly differentiates the two tasks. Tag completion evaluation uses only the semantics learnt from co-occurring tags. Thus, Tag completion becomes an appropriate task to evaluate the quality of semantics learnt by the CE model. It is worth stating that the two selected evaluation tasks are not to be confused with Audio Tag Classification task. In the later, music is associated with the relevant tags in one of two modes, binary mode and ranking mode. In both modes, the music content is considered which is not true for our evaluations. Indeed, semantic priming and tag completion are technically similar as they both operate on the same semantic space, but the difference in nature of the used data results in different evaluation settings. Furthermore, semantic priming, being most generic, is used for early stopping of the CE model training (see Section 2.3.) while tag completion only measures the performance of such semantics in this specific setting.

### 4.1 Dataset

Continuous track-tag relatedness ground-truth is rarely available for music tag collections as generating such information demands high-level of expertise and is extremely laborious. Fortunately, CAL500 and MSDSub provide the continuous track-tag relatedness ground truth alongside the binary relatedness information [Turnbull et al. 2007; Bertin-mahieux et al. 2011]. Thus, we employ these two datasets in this evaluation. As a result, each evaluation document from either dataset has two annotation versions, the binary one and the continuous one. In our experiments, we assume that an evaluation document has already been annotated with a group of existing tags based on its binary relatedness version. This assumption is based on the fact that it is impossible to split documents into coherent sets of tags automatically. The evaluation was conducted on the subsets of two datasets, i.e., 127 and 3,487 documents in CAL500 and MSDSub, respectively.

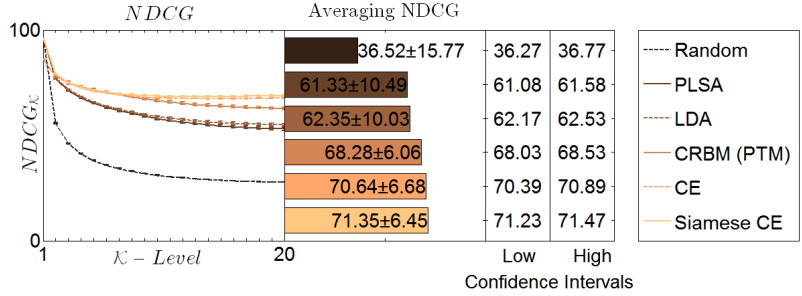

Figure 9: Normalized Discounted Cumulative Gain evaluation results using CAL500. NDCG values at different K levels with the standard error as well as the Averaging of NDCG up to 20 levels.

### 4.2 Evaluation

Given a tag in local context, all tags are ranked according to the predicted relevance in terms of contextualized semantics learnt by a model. This predicted tag ranking, $r$, is evaluated against the continuous relevance ground truth. For evaluation, we employ the *Normalized Discounted Cumulative Gain* (NDCG) measure [Järvelin and Kekäläinen 2002] across an entire retrieved and ranking list. Given $r[i]$ is the $i^{th}$ element in $r$, and $rel(r[i])$ is the ground-truth relevance value of tag $r[i]$; we measure *Discounted Cumulative Gain* (DCG) up to position $k$ as $DCG_k(r) = rel(r[1]) + \sum_{i=2}^{k} \frac{rel(r[i])}{log(i)}$. Similar to P@$k$, this measure is affected by the length of the document. Therefore, we normalize the DCG by using the best possible DCG for each evaluation document, i.e., the list $r_{ideal}$, the ideal ranking of tags derived from the ground truth given the query tag in context. Thus, the NDCG is measured

$$NDCG_k = \frac{DCG_k(r)}{DCG_k(r_{ideal})},$$

which assigns higher values to predicted lists with tag ranking closer to the ideal ranking. Changing parameter $k$, we achieve different $NDCG_k$ values. An Averaging NDCG up to $K$ is obtained by averaging all the $NDCG_k$ values for $k \leq K$.

### 4.3 Results

In our experiments, a document with $m$ tags results in $m$ separate queries for evaluation although they share the same local context. It is worth mentioning that an alternative setting could be aggregating $m$ tags into one query and rank all other tags according to their relatedness to the document query. However, the latter setting does not provide any better insight and may, in fact, cause lose in granuality.

Figure 9 shows the NDCG results on CAL500 including the Averaging $NDCG_k$ values at different $k$ levels with the standard error due to the three-fold CV as well as the averaging NDCG value up to $K = 20$. All models perform similarly at the small $k$ levels due to the high density underlying this dataset. However, predicting up to 20 reasonable tags appears to be a challenging problem given the fact that the $NDCG_k$ values drop sharply as $k$ increases. Nevertheless, the CE model generally outperforms all others especially for $k > 10$ as shown in Figure 9. The same results are confirmed by the averaging NDCG values up to $K = 20$. Overall, Siamese CE performs slightly better than CE due to the semantic distance learning.

Similarly, Figure 10 shows the NDCG results on MSDSub. Unlike the results on CAL500, the CE model yields the better performance even at small $k$ levels as shown in Figure 10. In MSDSub, the average document length is 3.2. Hence, the performance for $k > 3$ would clearly reflect the capability of a model in identifying those reasonable "negative" tags specified in the ground truth. The evidence seen in Figure 10 strongly suggests that the contextualized semantics learnt by the CE model leads to better performance with statistical significance (Student's t-test p-value< 0.01).

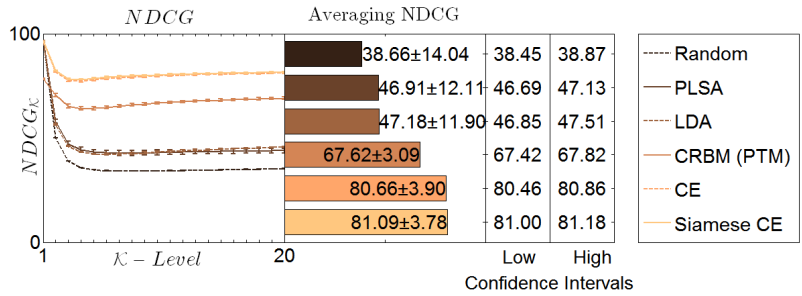

Figure 10: NDCG performance of different models on MSDSub.

In summary, the tag completion evaluation further demonstrates the effectiveness of the resultant contextualized semantics learnt by the CE model in capturing the intention of tags' use beyond the co-occurrence statistics. Results reported in two benchmark tasks suggest that the semantics learnt by the CE model is ready for use in various MIR applications. It is also worth stating that the CE model may be applied flexibly given the fact that in general, Siamese CE merely outperforms CE slightly on test data but performs considerably better than CE on training data (see the appendix for details). That is, CE would be employed for those MIR tasks that involve many unseen documents in training in order to reduce the training time (c.f. Table I). Otherwise, Siamese CE should be applied for retrieval purposes only.

## 5. DISCUSSION

Unlike previous approaches in semantics learning, the CE model learns distributed semantics without taking any particular MIR tasks into account. This should bring us closer to bridging the semantic gap encountered by various MIR tasks. By considering the local context, our approach leads to distributed multi-representations of a tag associated with different semantic contexts. This salient characteristic distinguishes our approach from others in learning music semantics from tags.

Previous work in music understanding [Turnbull et al. 2007; Miotto and Lanckriet 2012] focused on performance in one task, namely music annotation or auto-tagging. Indeed, good performance in mapping acoustic content to tags should assist music understanding. However, the resulting models are limited to the training tags only. Moreover, the reliance on acoustic content may result in inconsistent semantics since the contextualizing cues of a tag may have different patterns and are often mixed with other content components. Furthermore, blindly learning to annotate the data by following the given labels may result in biases towards the opinions of the labelers as well as overly fitting any noisy labels in the data [Sturm 2014]. We argue that these limitations can be overcome by using knowledge in the form of semantics learnt from crowd-source tag collections as demonstrated in the tag completion task.

In general, *expertise-based* semantics is transferable across MIR tasks, including a) *attributes listing*: the attributes of a tag are manually listed and comparisons are made on the attribute level; b) *ontology*: concepts are manually associated [Kim et al. 2008]; and c) *knowledge base*: first-order logic rules governing relatedness of tags are maintained [Wang et al. 2010]. Moreover, ontologies have been employed as a categorizing scheme and as a filtering step for tags that happen to be in a dictionary [Cantador et al. 2011]. In comparison to expert-based semantics, CE learnt semantics is less interpretable unless additional information is available.

Nevertheless, expert-based semantics incurs intensive handcrafted work and suffers from an intrinsic difficulty in quantifying relatedness. These difficulties become more severe in the presence of OOV tags and contexts to consider. In contrast, our approach effortlessly leads to the contextualized semantics and is capable of dealing with OOV tags. Thanks to the limited human intervention in establishing and maintaining CE semantics, which are automatically learnt from public tag collections, we firmly believe that it is transferable without biases; and, hence, is greatly applicable in various MIR tasks.

Sometimes semantics may be obtained from multiple sources, including artist/track information, users' playlists preferences, tagging information and music production information [Mandel et al. 2011; Weston et al. 2011]. Furthermore, personal intention has been investigated in previous studies [Mika 2005; Mandel et al. 2011]. In such work, the personal meaning of a tag is inferred by analyzing user's tagging activities. Other information sources might include the manual categorization of documents into semantic classes [Font et al. 2014; Font 2015, pp.67–90], which allows for better within-class tag similarity estimation once the proper class of a test document is identified. The motivation behind such methods is the construction of rich, transferrable and often personalized semantics. Aside from the high labor cost, there is no guarantee that the multiple semantics sources are complementary and consistent as integrated semantics. Moreover, finding out possible contradictions between multiple sources may require additional human intervention and more training data. Incomplete and noisy sources further aggravate the semantics fusion. In contrast, our approach does not rely on such information and explores the objective meaning of tags instead. Such objective semantics might easily be used in personalization systems later when coupled with users' information sources or profiles. Thus, we believe that learning semantics from co-occurring tags, a single informative source, is sufficient and justifiable.

Throughout this paper, we focused on the contextualized relatedness models. However, global relatedness models have also been used in semantics acquisition due to their simplicity and ease of use. For instance, a specific PCA model is proposed in [Lebret et al. 2013] that improved the accuracy of movie review sentiment evaluation. Another successful model is presented in [Mandel et al. 2011] where an information theoretic method is used to produce a smoothed representation of a training document which is subsequently applied in music annotation. Other examples include [Mika 2005], where a tag is represented via the information of who used it, and [Font et al. 2014], where the similarity of tags within each "class" is established via aggregation (c.f. Section 2.1). We acknowledge the usefulness of such models in the music annotation tasks. However, there is evidence that music annotations are contextualized and that meaning cannot be uniquely identified without considering the companion tags as described in Section 1. Similarly, syntactic models learnt from large text corpora [Mikolov et al. 2013] yield a single representation for each word. Apart from the fact that these models encode global relatedness only, musical tags are not always in single word form and may contain phrases, e.g. "acoustic guitar", symbols, e.g. "90s", and abbreviations, e.g. "r'n'b". On one hand, it is infeasible to apply these models without domain adaptations into MIR. On the other, adapting such models demands semantics governed by linguistic rules that are absent in tags.

## 6. FUTURE WORKS AND CONCLUSIONS

Exploring intrinsic semantics underlying tags demands a powerful context model that can model the contextual information effectively. PTMs used in the CE model provide a powerful yet generic tool for information aggregation from documents. Unfortunately, such models lack interpretability of the modeled semantics and hence cause the CE model to suffer from the same limitation. For a specific application, however, there is often alternative contextual information and relevant modeling techniques, e.g., the labeled LDA [Ramage et al. 2009] trained with meaningful topic labels such as genre or instrument types. In this situation, it is straight-forward to incorporate such context representations into the CE model proposed in Section 2. An extension of the CE model by using different context representations may facilitate some MIR tasks that require the self-defined context by users, e.g., a playlist of music in a specific style or mood. Moreover, a context modeling the user's behavior can be inferred and issued along with the user's query tags for semantically-coherent query expansion. In fact, the success of the CE model in the Tag completion task hints its suitability for proper query expansion where we explicitly add semantically related keywords to a user's query in order to better describe the user's need in their specific scenario and improve the results of a retrieval task.

Given proper music and/or artist representation within the semantics space, e.g. using aggregation of their tags, it becomes straightforward to *prime* music tracks or artists given user's query concepts. For instance, we can semantically describe a music track as the centroid and the spread statistics of its tags within the semantic space; effectively achieving a semantically descriptive location for each track within the semantic space. As a result, not only a user can query a system for musical content, but also a semantic-level similarity measure between tracks is achieved and can be used for music discovery. Moreover, it becomes easy to annotate music or measure the similarity of artists' work. For example, using a mapping function from the content to the semantic space would facilitate annotating music with known and OOV tags. Such function can also be used to query by example systems.

Finally, the CE model requires the tag features and the BoW output representation. This requirement may prevent the CE model from being applied to these datasets of a large vocabulary, e.g., the complete Million Song Dataset (MSD), due to the high computational burden. There are some technique that could potentially overcome this limitation by generating parsimonious representations, e.g., PCA, compressed sensing [Hsu et al. 2009] and filtering techniques [Cantador et al. 2011]. However, such techniques are still under investigation in our ongoing work.

In conclusion, we presented a comprehensive argument for the suitability of contextualized music semantic representations from co-occurring tags in the music domain. The contextualized semantics learnt by CE approach significantly outperforms several state-of-the-art semantic learning methods as suggested in our semantic priming and tag completion evaluation. Moreover, the semantics learnt by the CE model properly deals with the sparse BoW situation as well as the OOV tags. While the work presented in this paper is only regarding the contextualized semantic learning from tags and the evaluation was conducted on two generic benchmark tasks, the semantics learned by our approach can be potentially applied to various MIR tasks. As the approach described in this paper is a generic approach in learning semantics from any types of tags or descriptive terms regardless of media type, we plan to apply our approach to other domains in our ongoing work.

# REFERENCES

Yoshua Bengio, Pascal Lamblin, Dan Popovici, and Hugo Larochelle. 2007. Greedy layer-wise training of deep networks. In Bernhard Schölkopf, John C. Platt, and Tim Hoffman (Eds.). *Advances in Neural Information Processing Systems 19 (NIPS'06)*. Vancouver, BC, Canada., 153–160.

Thierry Bertin-Mahieux, Douglas Eck, and Michael Mandel. 2010. Automatic tagging of audio: the state-of-the-art. In Wenwu Wang (Ed.). *Machine Audition: Principles, Algorithms and Systems*. Hershey, PA, USA.: IGI Publishing, 334–352.

Thierry Bertin-mahieux, Daniel P.W. Ellis, Brian Whitman, and Paul Lamere. 2011. The million song dataset. In *Proceedings of the 12th International Society for Music Information Retrieval Conference (ISMIR 2011)*. Miami, FL, USA., 591–596.

David Blei, Andrew Ng, and Michael I. Jordan. 2003. Latent Dirichlet allocation. *J. Mach. Learn. Res.* 3 (2003), 993–1022.

Léon Bottou. 2012. Stochastic gradient descent tricks. In Grégoire Montavon, Geneviève Orr, and Klaus-Robert Müller (Eds.). *Neural Networks, Tricks of the Trade, Reloaded*. Springer-Verlag Berlin Heidelberg, 430–445.

Jane Bromley et al. 1993. Signature verification using a "Siamese" time delay neural network. *Int. J. Pattern Recognit. Artif. Intell.* 7, 4 (August 1993), 669–688.

Iván Cantador, Ioannis Konstas, and Joemon M. Jose. 2011. Categorising social tags to improve folksonomy-based recommendations. *Web Semant. Sci. Serv. Agents World Wide Web.* 9, 1 (March 2011), 1–15.

Jacob Cohen. 1977. *Statistical Power Analysis for the Behavioral Sciences* Revised Ed., New York, NY, USA.: Academic Press, INC. LTD.

Scott Deerwester, Susan T. Dumais, George W. Furnas, Thomas K. Landauer, and Richard Harshman. 1990. Indexing by latent semantic analysis. *J. Am. Soc. Inf. Sci.* 41, 6 (1990), 391–407.

Frederic Font. 2015. *Tag Recommendation using Folksonomy Information for Online Sound Sharing Platforms*. Thesis. Universitat Pompeu Fabra.

Frederic Font, Joan Serrà, and Xavier Serra. 2014. Class-based tag recommendation and user-based evaluation in online audio clip sharing. *Knowledge-Based Syst.* 67 (2014), 131–142.

Zellig S. Harris. 1954. Distributional structure. *WORD-Journal Int. Linguist. Assoc.* 10, 2-3 (1954), 146–162.

Thomas Hofmann. 1999. Probabilistic latent semantic indexing. In *Proceedings of the 22nd ACM SIGIR conference on Research and development in information retrieval*. New York, NY, USA.: ACM Press, 50–57.

Daniel Hsu, Sham M. Kakade, John Langford, and Tong Zhang. 2009. Multi-Label prediction via compressed sensing. In Yoshua Bengio, Dale Schuurmans, John D. Lafferty, Christopher K. I. Williams, and Aron Culotta (Eds.). *Advances in Neural Information Processing Systems 22 (NIPS'09)*. Vancouver, BC, Canada., 772–780.

Ignacio Iacobacci, Mohammad Taher Pilehvar, and Roberto Navigli. 2015. SensEmbed: Learning Sense Embeddings for Word and Relational Similarity. In *Proceedings of the 53rd Annual Meeting of the Association for Computational Linguistics (ACL 2015)*. Beijing, China, 95–105.

Kalervo Järvelin and Jaana Kekäläinen. 2002. Cumulated gain-based evaluation of IR techniques. *ACM Trans. Inf. Syst.* 20, 4 (2002), 422–446.

Hak Lae Kim, Simon Scerri, John G. Breslin, Stefan Decker, and Hong Gee Kim. 2008. The state of the art in tag ontologies: A semantic model for tagging and folksonomies. In *Dublin Core and Metadata Applications*. Berlin, Germany.: The Dublin Core Metadata Initiative., 128–137.

Edith Law, Luis von Ahn, Roger B. Dannenberg, and Mike Crawford. 2007. TagATune: A game for music and sound annotation. In *Proceedings of the 8th International Society for Music Information Retrieval Conference (ISMIR 2007)*. Vienna, Austria., 361–364.

Edith Law, Burr Settles, and Tom Mitchell. 2010. Learning to tag from open vocabulary labels. In José L. Balcázar, Francesco Bonchi, Aristides Gionis, and Michèle Sebag (Eds.). *Proceedings of the 2010 European conference on Machine learning and knowledge discovery in databases (EMCL PKDD 2010)*. Barcelona, Spain.: Springer Berlin Heidelberg, 211–226.

Edith Law, Kris West, Michael Mandel, Mert Bay, and J. Stephen Downie. 2009. Evaluation of algorithms using games: the case of music tagging. In *10th International Society for Music Information Retrieval Conference (ISMIR 2009)*. Kobe, Japan., 387–392.

Rémi Lebret, Joël Legrand, and Ronan Collobert. 2013. Is deep learning really necessary for word embeddings? In *Advances in Neural Information Processing Systems 26 (NIPS'13), Deep Learning Workshop*. Lake Tahoe, NV, USA.

Mark Levy and Mark Sandler. 2008. Learning latent semantic models for music from social tags. *J. New Music Res.* 37, 2 (June 2008), 137–150.

Michael S. Lew, Nicu Sebe, Chabane Djeraba, and Ramesh Jain. 2006. Content-based multimedia information retrieval: State of the art and challenges. *ACM Trans. Multimed. Comput. Commun. Appl.* 2, 1 (2006), 1–19.

Kevin Lund and Curt Burgess. 1996. Producing high-dimensional semantic spaces from lexical co-occurrence. *Behav. Res. Methods, Instruments, Comput.* 28, 2 (1996), 203–208.

Laurens van der Maaten and Geoffrey Hinton. 2008. Visualizing data using t-SNE. *J. Mach. Learn. Res.* 9 (2008), 2579–2605.

Michael I. Mandel et al. 2011. Contextual tag inference. *ACM Trans. Multimed. Comput. Commun. Appl.* 7S, 1, Article 32 (2011), 18 pages.

Christopher D. Manning, Prabhakar Raghavan, and Hinrich Schütze. 2009. *An Introduction to Information Retrieval*, Cambridge, England: Cambridge University Press, Online edition. Retreived from http://www.informationretrieval.org.

Benjamin Markines, Ciro Cattuto, Filippo Menczer, Dominik Benz, Andreas Hotho, and Gerd Stumme. 2009. Evaluating similarity measures for emergent semantics of social tagging. In *Proceedings of the 18th International Conference on World Wide Web (WWW'09)*. Madrid, Spain.: ACM Press, 641–650.

Gonçalo Marques, Marcos Aurélio Domingues, Thibault Langlois, and Fabien Gouyon. 2011. Three current issues in music autotagging. In *Proceedings of the 12th International Society for Music Information Retrieval Conference (ISMIR 2011)*. Miami, FL, USA., 795–800.

Peter Mika. 2005. Ontologies Are Us: A Unified Model of Social Networks and Semantics. In Yolanda Gil, Enrico Motta, V. Richard Benjamins, and Mark A. Musen (Eds.). *The Semantic Web – ISWC 2005*. Galway, Ireland.: Springer Berlin Heidelberg, 522–536.

Tomas Mikolov, Greg Corrado, Kai Chen, and Jeffrey Dean. 2013. Efficient estimation of word representations in vector space. In *International Conference on Learning Representations (ICLR)*. Scottsdale, AZ, USA.

Riccardo Miotto and Gert Lanckriet. 2012. A generative context model for semantic music annotation and retrieval. *IEEE Trans. Audio. Speech. Lang. Processing* 20, 4 (2012), 1096–1108.

Rajat Raina, Anand Madhavan, and Andrew Y. Ng. 2009. Large-scale deep unsupervised learning using graphics processors. In *Proceedings of the 26th Annual International Conference on Machine Learning (ICML '09)*. 873–880.

Daniel Ramage, David Hall, Ramesh Nallapati, and Christopher D. Manning. 2009. Labeled LDA: A supervised topic model for credit attribution in multi-labeled corpora. In Philip Koehn and Rada Mihalcea (Eds.). *Proceedings of the 2009 Conference on Empirical Methods in Natural Language Processing: Volume 1*. Singapore: ACM, 248–256.

Ubai Sandouk and Ke Chen. 2016. Learning contextualized semantics from co-occurring terms via a Siamese architecture. *Neural Networks* 76 (April 2016), 65–96.

Markus Schedl, Arthur Flexer, and Julián Urbano. 2013. The neglected user in music information retrieval research. *J. Intell. Inf. Syst.* 41, 3 (2013), 523–539.

Xavier Serra et al. 2013. *Roadmap for Music Information ReSearch (MIReS)* Geoffroy Peeters (Ed.)., The MIReS Concortium.

Amit Singhal. 2001. Modern information retrieval: A brief overview. *IEEE Data Eng. Bull.* 24 (2001), 35–43.

Arnold W.M. Smeulders, Marcel Worring, Simone Santini, Amarnath Gupta, and Ramesh Jain. 2000. Content-based image retrieval at the end of the early years. *IEEE Trans. Pattern Anal. Mach. Intell.* 22, 12 (2000), 1349–1380.

Mohamed Sordo. 2011. *Semantic Annotation of Music Collections: A Computational Approach*. Ph.D. Dissertation. Universitat Pompeu Fabra. Barcelona, Spain.

Bob L. Sturm. 2014. The state of the art ten years after a state of the art: Future research in music information retrieval. *J. New Music Res.* 43, 2 (May 2014), 147–172.

Yee Whye Teh, Michael I. Jordan, Matthew J. Beal, and David M. Blei. 2006. Hierarchical Dirichlet processes. *J. Am. Stat. Assoc.* 101, 476 (2006).

Douglas Turnbull, Luke Barrington, and Gert Lanckriet. 2008. Five approaches to collecting tags for music. In *Proceedings of the 9th International Society for Music Information Retrieval Conference (ISMIR 2008)*. Philadelphia, PA, USA.

Douglas Turnbull, Luke Barrington, David Torres, and Gert Lanckriet. 2007. Towards musical query-by-semantic-description using the CAL500 data set. In *Proceedings of the 30th annual international ACM SIGIR conference on Research and development in information retrieval (SIGIR'07)*. Amsterdam, Netherlands.: ACM Press, 439–446.

Jun Wang, Xavier Anguera, Xiaoou Chen, and Deshun Yang. 2010. Enriching music mood annotation by semantic association reasoning. In *2010 IEEE International Conference on Multimedia and Expo*. Singapore.: IEEE, 1445–1450.

Jason Weston, Samy Bengio, and Philippe Hamel. 2011. Multi-tasking with joint semantic spaces for large-scale music annotation and retrieval. *J. New Music Res.* 40, 4 (2011), 337–348.

# Appendix
# Learning Contextualized Music Semantics from Tags via a Siamese Neural Network

UBAI SANDOUK and KE CHEN, University of Manchester

In this appendix, we report the semantic priming results achieved by different models in learning semantics from tags on training subsets of three music datasets described in Section 3.1 in the main text. The same experimental protocol and notation specified in Section 3.5 is also used here.

Figure A.1 shows the semantic priming results on CAL500. The dataset is highly dense and predicting the first few tags (small $k$ values) appears relatively easy. However, the CE model seems to be over-fitting in Siamese training as observed from the Precision-Recall curves despite the use of a validation procedure during training. By comparison to the results on the test subsets shown in Figure 4 in the main text, it is seen from Figure A.1 that Siamese CE considerably outperforms all other models including CE on the training subsets while it performs only slightly better than CE where the confidence intervals reveal its statistically significant advantage. Moreover, the effective difference between Siamese CE over CRBM(PTM) is 16.32 in MAP and 26.5 in AUC, and the effective difference between Siamese CE and CE is 7.43 in MAP and 8.98 in AUC. These results suggest a clear advantage for Siamese CE over other models in all recall ranges on the training subsets.

Figure A.2 shows the semantic priming results on MagTag5K. It is evident from Figure A.2 that Siamese CE also outperforms other models including CE with statistical significance although the gain is smaller than that on CAL500. We also observe that LDA and PLSA perform similarly and achieve little improvement over the Random model. The over-fitting is less on this dataset than CAL500 as the Siamese CE not only outperforms CE but also reduces variance of results amongst different CV folds. Finally, effective difference between Siamese CE and CRBM(PTM) is 15.51 in MAP and 24.61 in AUC, and the effective difference between Siamese CE and CE is 1.62 in MAP and 3.14 in AUC.

Figure A.3 shows the semantic priming results on MSDSub. Due to the very short document length, i.e., 3.2 tags on average, in MSDSub, the $P@k$ performance drops quickly after $k = 3$ for all the models. The sparse nature of this dataset also causes Siamese CE to slightly outperform CE. Nevertheless, our model performs very well in the high recall range while LDA and PLSA perform merely slightly better than the Random model. The large vocabulary and the sparsity of this dataset account for the deteriorated performance of other models but affect our model very little. In addition, the effective difference between Siamese CE and CRBM(PTM) reaches 56.1 in MAP and 90.31 in AUC, while the effective difference between Siamese CE and CE is 1.97 in MAP and 1.82 in AUC. Due to the size of the dataset, the effective difference in MAP is larger than that observed on MagTag5K. Once again, we emphasize that the seemingly modest advantage of Siamese CE is difficult to obtain but useful for real world MIR applications.

In summary, we report the experimental results on the training subsets to present a complete picture on the contextualized semantics learning by the CE model. While the generalization capability is very important from a machine learning perspective, the good performance on training data may be helpful in information retrieval. Our experimental results on training data reported above clearly shows that Siamese CE outperforms all other models including CE, in particular, in the high recall range. Thus, we firmly believe that Siamese CE offers an additional gain for various MIR applications that benefit from identifying *all* related tags.

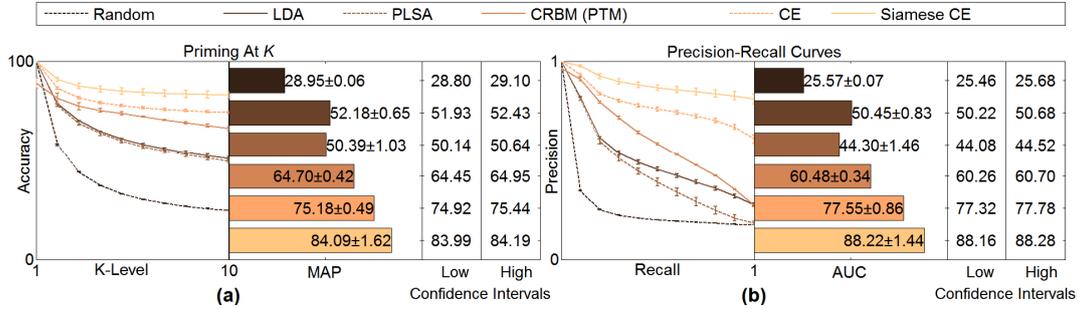

Figure A.1: Semantic priming performance of different models on the training subsets of CAL500.

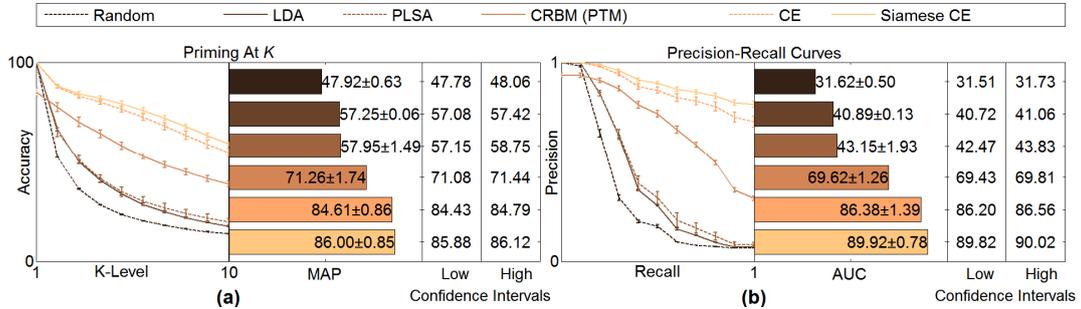

Figure A.2: Semantic priming performance of different models on the training subsets of MagTag5K.

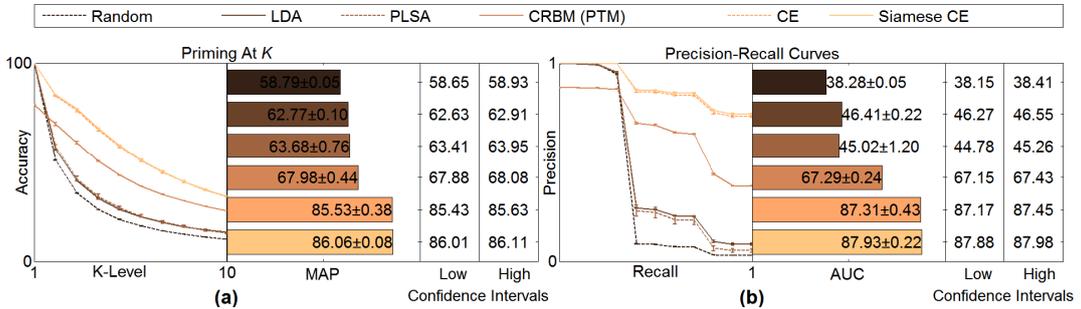

Figure A.3: Semantic priming performance of different models on the training subsets of MSDSub.